\documentclass[12pt,letterpaper]{article}
\usepackage[a4paper, total={7in, 10in}]{geometry}

\usepackage{helvet}
\usepackage{authblk}
\usepackage{array}
\usepackage{mdwmath}
\usepackage{mdwtab}
\usepackage{eqparbox}
\usepackage{url}
\usepackage{multirow}
\usepackage{bbding}
\usepackage{booktabs,array}
\usepackage{threeparttable}
\usepackage{tabularx}
\usepackage{booktabs}
\usepackage{makecell}
\usepackage{amsfonts}
\usepackage{amsmath}
\usepackage{amssymb}
\usepackage{verbatimbox}
\usepackage{framed}
\usepackage{graphicx}
\usepackage{hyperref}

\makeatletter
\renewcommand{\maketitle}{\bgroup\setlength{\parindent}{0pt}
\begin{flushleft}
  \textbf{\@title}
  
  \@author
\end{flushleft}\egroup}
\makeatother

\title{Exposing Image Splicing Traces in Scientific Publications via Uncertainty-guided Refinement}
\date{}

\author[1]{Xun Lin}
\author[1]{Wenzhong Tang}
\author[1]{Haoran Wang}
\author[1]{Yizhong Liu}
\author[2]{Yakun Ju}
\author[1,*]{Shuai Wang}
\author[3,*]{Zitong Yu}

\affil[1]{Beihang University, Beijing, China}
\affil[2]{Ocean University of China, Qingdao, China}
\affil[3]{Great Bay University, Dongguan, China}
\affil[*]{Correspondence: wangshuai@buaa.edu.cn, yuzitong@gbu.edu.cn}

\usepackage[super,comma,sort&compress]{natbib}

\newcommand{\figref}[1]{Fig.~\ref{#1}}
\newcommand{\tabref}[1]{Table~\ref{#1}}
\newcommand{\eqnref}[1]{Eq.~(\ref{#1})}

\newcommand{\model}{URN}
\newcommand{\ie}{\textit{i.e., }}
\newcommand{\eg}{\textit{e.g., }}
\newcolumntype{P}[1]{>{\centering\arraybackslash}p{#1}}
\newcolumntype{M}[1]{>{\centering\arraybackslash}m{#1}}

\begin{document}

\maketitle

\section*{Summary}
% TODO: 引用
Recently, a surge in scientific publications suspected of image manipulation has led to numerous retractions, bringing the issue of image integrity into sharp focus. 
Although research on forensic detectors for image plagiarism and image synthesis exists, the detection of image splicing traces in scientific publications remains unexplored. 
Compared to image duplication and synthesis, image splicing detection is more challenging due to the lack of reference images and the typically small tampered areas. 
Furthermore, disruptive factors in scientific images, such as artifacts from digital compression, abnormal patterns, and noise from physical operations, present misleading features like splicing traces, significantly increases the difficulty of this task. Moreover, the scarcity of high-quality datasets of spliced scientific images limits potential advancements. In this work, we propose an Uncertainty-guided Refinement Network (URN) to mitigate the impact of these disruptive factors. Our URN can explicitly suppress the propagation of unreliable information flow caused by disruptive factors between regions, thus obtaining robust splicing features. Additionally, the URN is designed to concentrate improvements in uncertain prediction areas during the decoding phase. We also construct a dataset for image splicing detection (SciSp) containing 1,290 spliced images. Compared to existing datasets, SciSp includes the largest number of spliced images and the most diverse sources. Comprehensive experiments conducted on three benchmark datasets demonstrate the superiority of our approach. We also validate the URN's generalisability in resisting cross-dataset domain shifts and its robustness against various post-processing techniques, including advanced deep-learning-based inpainting.

\section*{Keywords}

Scientific integrity, image splicing detection, convolution neural network, and uncertainty

\section*{Introduction}

\begin{figure}[t]
	\setlength{\abovecaptionskip}{4pt}
	\setlength{\belowcaptionskip}{-12pt}
	\centering 
	\includegraphics[width=0.7\textwidth]{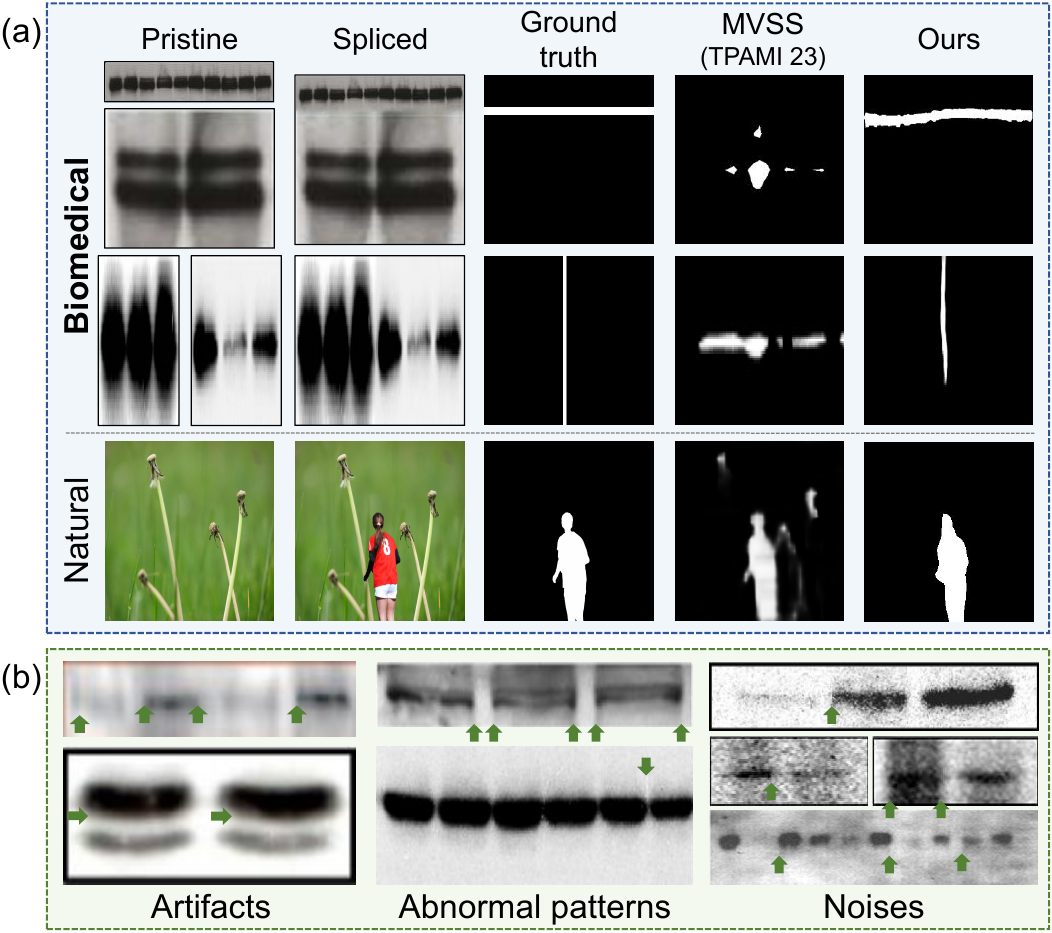}
	\caption{(a) Difference between spliced scientific (top two rows) and natural image (bottom one row). Examples of splicing detection results on the two kinds of images predicted by our method and a SoTA natural image manipulation detection method, \textit{i.e.} MVSS \cite{mvss}. (b) Disruptive factors such as artifacts, abnormal patterns, and noises in scientific images. The green arrows indicate regions interfered with disruptive factors where the SoTA manipulation detectors designed for natural images tend to give false alarms.} 
	\label{fig:simple} 
\end{figure}

In scientific publications, erroneous conclusions and irreproducible experimental results caused by inappropriate image manipulation pose significant threats to the scholarly community, especially for biomedical research \cite{ross2004}.
This issue can mislead academic peers, lead to a serious waste of resources, and decrease public confidence in scientific research \cite{bik2016}.
Scientific images, as essential components of scientific publications, always serve to demonstrate experimental results \cite{miura2021reproducible}. As shown in \figref{fig:simple}(a), malicious researchers can manipulate the images to fabricate experimental conclusions or conceal unfavorable results \cite{auto_cell}, which pose significant threats to the scholarly community \cite{ross2004}. 

With advancements in deep learning techniques and the availability of scientific image forensics datasets \cite{biofors,rsiid,binder}, notable progresses have been made in detecting duplication \cite{binder,sila,monet}, synthesis \cite{deepfake1,deepfake2} and copy-move manipulation \cite{sila}. 
However, splicing, as one of the most commonly used manipulations for scientific images, is more challenging to detect but receives less attention. 
Compared to duplication detection and copy-move detection, splicing detection lacks sufficient references for comparison and retrieval. Furthermore, image synthesis, as opposed to image splicing, tends to generate more detectable forgery cues because image synthesis generally leads to global forgeries, whereas image splicing results in localized forgeries.
To the best of our knowledge, no method has been specifically proposed for scientific image splicing detection. 

However, the detection of manipulated (including spliced) pixels in natural images captured by digital cameras has been extensively studied, and several standard datasets are proposed for this purpose \cite{luisa_review}.
In the early years, with the boost of deep learning techniques, GSR \cite{gsr}, and RRU \cite{rru} introduce end-to-end Convolutional Neural Networks (CNNs) for natural image manipulation detection.
The effectiveness of self-attention mechanisms in this task has also been discussed by TransForensics \cite{transforensics} and PSCC-Net \cite{pscc}.
Analyzing intrinsic statistics to explore traces of image forgeries also gains progress \cite{catnet,catnet2,objectformer}.
Towards more generalized detection, many methods employ noise-sensitive filters \cite{rgbn,mantra,mvss,emt} or self-supervised learning \cite{noiseprint,trufor} to suppress semantic information and analyze noise inconsistency of images.
More recently, Hifi \cite{hifi} expands this task, proposing fine-grained image forgery detection with hierarchical labels. It seems to be a feasible way to train the aforementioned methods on scientific datasets. Yet, existing natural image manipulation detectors, whether proposed for splicing detection \cite{rru} or general purposes \cite{trufor,catnet2,hifi,gsr,pscc,catnet2,mvss,mantra}, cannot achieve satisfactory performance on scientific datasets \cite{biofors}. 

We attribute the failure to two primary reasons: (1) the prevalence of more disruptive factors in scientific images and (2) the limited number of spliced images for training. 
Such disruptive factors can mislead detection methods, resulting in false alarms and incomplete predictions. These factors, illustrated in \figref{fig:simple}(b), are enumerated as follows: 
%These factors are enumerated as follows.  
\begin{itemize}
\item \textbf{Artifacts.} Scientific images always undergo multiple types of non-malicious degradation by authors and publishers. Commonly used degradations, \textit{e.g.} JPEG compression \cite{csvt-compress}, can decrease tampering traces and bring more artifacts \cite{osn}.

\item \textbf{Abnormal patterns.} During experimental processes, operational mistakes (\eg improper reagent ratios and gel rupture) can result in abnormal patterns, which tend to be confused with spicing traces.

\item \textbf{Noise.} Malfunctions and degradations of imaging devices or even simple operator errors can introduce a large amount of noise.
\end{itemize}

Furthermore, researchers with malicious intent can use advanced image editing tools \cite{densefcn} such as Photoshop\footnote{https://www.adobe.com/products/photoshop.html} or advanced AI-based generative models, \eg Generative Adversarial Networks (GAN) \cite{lama} and diffusion models \cite{diffusion}, to post-process the spliced images. 
These post-processing approaches are able to reduce visible clues of tampering \cite{csvt-post}, making it difficult to distinguish between the natural boundaries and splicing traces. 

\begin{figure*}[t]
	\setlength{\abovecaptionskip}{4pt}
	\setlength{\belowcaptionskip}{-9pt}
	\centering 
	\includegraphics[width=1.0\textwidth]{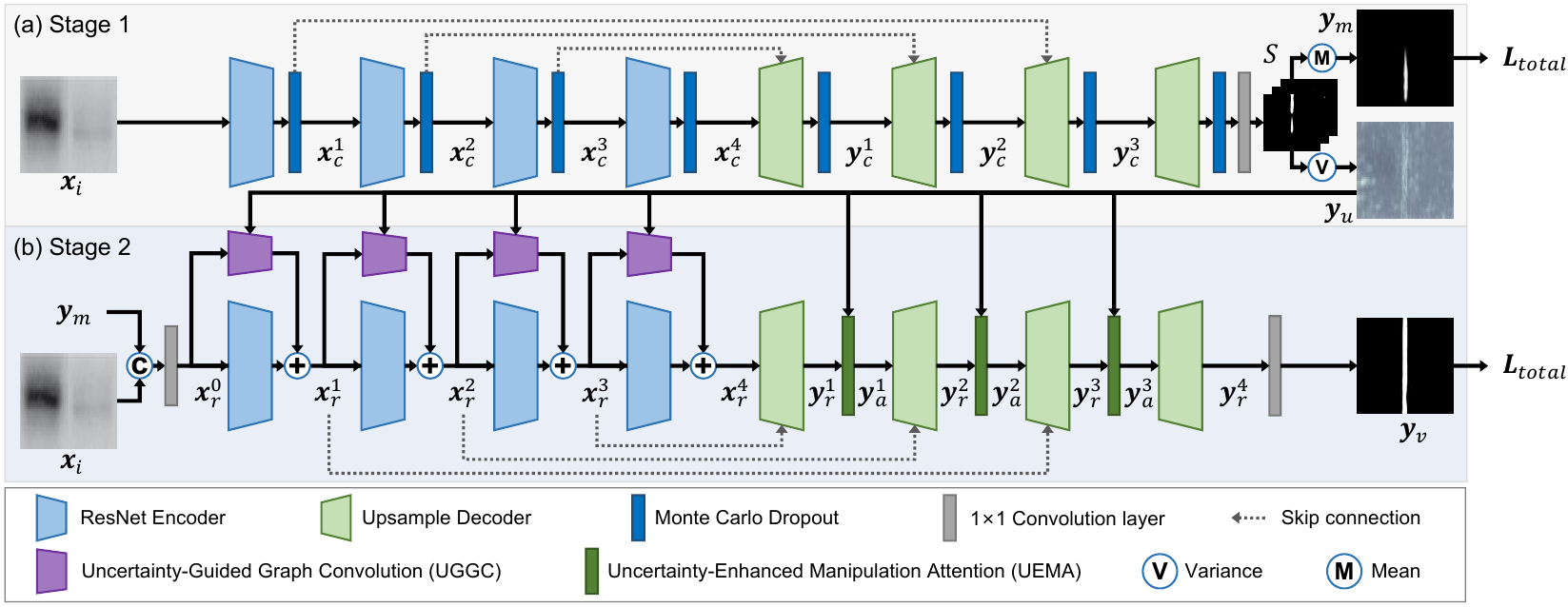}
	\caption{Overall structure of the proposed \model. It consists of two stages: (a) the stage-1 network makes coarse predictions of spliced regions and estimates the pixel-level uncertainty about coarse predictions, and (b) the stage-2 uncertainty-guided refinement network can refine coarse predictions under the guidance of uncertainty information.} 
	\label{fig:overall} 
\end{figure*}

To address such issues, we aim to design a robust framework that can resist disruptive factors and unknown post-processing approaches. As shown in \figref{fig:overall}, we propose a two-stage Uncertainty-guided Refinement Network (\model). 
In stage one, our \model~recognizes uncertainly predicted regions affected by disruptive factors using Monte Carlo Dropout (MCD). In the second stage, it integrates uncertainty to perform refinement. 
To fully exploit uncertainty information, we proposed Uncertainty-Guided Graph Convolution (UGGC) modules. 
The UGGC limits the unreliable information flow from uncertain regions but guides them to receive information from nearby confident regions. 
This ensures that the uncertain nodes can be gradually refined with less interference. In addition, we propose Uncertainty-Enhanced Manipulation Attention (UEMA) modules, 
which can focus on ambiguously predicted regions with high uncertainty and help distinguish spliced and pristine areas. 
By combining UGGC and UEMA, the performance and robustness of our URN can be further improved. 

Besides the effective methodologies, high-quality datasets are equally important for this task. 
Existing works, \ie RSIID \cite{rsiid}, BINDER \cite{binder}, and SYB \cite{syb}, automate the forgery of scientific images to generate large-scale forensics datasets, which are designed for the detection of manipulation, duplication, and synthesis, respectively.
To prevent the negative effects brought by the domain gap between automatically generated images and real-world ones, Biofors \cite{biofors} extracts 47,805 scientific images from papers published in \textit{PLOS ONE} 2013, including 1,741 manipulated images with pixel-level labels according to raw annotations provided by academic experts \cite{bik2016}.
It can be seen in \tabref{tab:dataset} that the quantity of spliced images remains insufficient, especially in Biofors which only contains 181 blot/gel images. 
The lack of high-quality datasets for scientific splicing detection hinders the progress of this field.
Methods trained on these datasets tend to have limited robustness and generality, which cannot be directly applied in real-world applications.
Sufficient datasets are available for the detection of scientific image duplication and copy-move manipulation. Benefiting from these datasets, significant progress has been made in these two tasks \cite{monet,sila}. 
However, the lack of high-quality datasets for scientific splicing detection hinders the progress of this field.
It can be seen in \tabref{tab:dataset} that the quantity of spliced images remains insufficient, especially in Biofors which only contains 181 blot/gel images. The lack of diversity in splicing approaches and image sources also hampers the training of a model with inspiring performance.

\begin{table}[t]
    \centering
    \caption{Details of datasets for scientific image splicing detection. All these datasets are used for evaluation. Names of our proposed sub-datasets are in bold.}
    \setlength{\tabcolsep}{14pt}
    \renewcommand{\arraystretch}{1.0}
    \label{tab:dataset}
    \resizebox*{1 \linewidth}{!}{
        \begin{tabular}{ccccccc}
            \toprule[1.3pt]
            \textbf{Name} & \begin{tabular}[c]{@{}c@{}}\textbf{Spliced} \\ \textbf{Images}\end{tabular} & \textbf{Type} & \begin{tabular}[c]{@{}c@{}}\textbf{Splicing} \\ \textbf{Approaches}\end{tabular} & \textbf{Sources} & \textbf{Acquisition}\\
            \midrule[1.3pt]
            Biofors \cite{biofors} & 181  & Blot/Gel & 3 & Found in Publications & from PLOS One\\
            RSIID \cite{rsiid} & 880  & Microscopy & 1 & Spliced by Algorithm & from 3 Public Datasets\\
            %			\rowcolor[HTML]{CCCCCC}
            \textbf{SciSp-C} & 290 & Blot/Gel & 3 & Found in Publications & from 39 Journals\\
            %			\rowcolor[HTML]{CCCCCC}
            \textbf{SciSp-H} & 1,000 & Blot/Gel & 5 & Manually Spliced & from PLOS One \\ \bottomrule[1.3pt]
        \end{tabular}
    }
\end{table}

Therefore, we propose a dataset for Scientific image Splicing detection (SciSp), a diverse and challenging dataset surpassing others in size, source diversity, and splicing complexity.
SciSp consists of two subsets: one is collected from public comments on the authoritative post-publication peer review platform PubPeer \footnote{https://pubpeer.com/}, while the other one comprises images manually manipulated with Photoshop.
In conclusion, our contributions can be summarized as follows:
\begin{itemize}
	\setlength{\topmargin}{0pt}
	\setlength{\itemsep}{0em}
	\setlength{\parsep}{0pt}
	\item We are the first to propose an end-to-end deep learning-based framework for scientific image splicing detection. Our Uncertainty-guided Refinement Network (\model) can resist disruptive factors in scientific images by fully using uncertainties to refine coarse predictions.
	\item We propose two modules, \ie the Uncertainty-Guided Graph Convolution (UGGC) and Uncertainty-Enhanced Manipulation Attention (UEMA) to help \model~integrate uncertainty information. UGGC is able to extract robust features against disruptive factors during encoding, while UEMA focuses on the refinement of uncertainly predicted areas in the decoding phase.
	\item We introduce a new dataset for Scientific image Splicing (SciSp) detection. Compared with existing datasets, SciSp has the largest number of spliced images from diverse sources and various splicing approaches.
	\item We conduct comprehensive experiments on four benchmarks, demonstrating the efficacy of the proposed method in both pixel-level and image-level detection.
\end{itemize}

\section*{Results}
% TODO: In this section
We use the F1-score (F1) and Matthews Correlation Coefficient (MCC) \cite{mcc} for pixel-level evaluation and adopt the Area Under the receiver operating characteristic Curve (AUC) and accuracy (Acc) for image-level evaluation.

\subsection*{Scientific image splicing detection performance}
To the best of our knowledge, no deep learning-based method was proposed for this task. Therefore, we adopt nine SoTA image manipulation or splicing detection methods designed for natural images. Note that all selected image manipulation methods can be used for splicing detection. These methods drop into two categories: (1) noise-sensitive methods, including ManTra \cite{mantra}, MVSS \cite{mvss} and TruFor \cite{trufor}, (2) noise-insensitive methods, \ie RRU-Net \cite{rru}, GSR-Net \cite{gsr}, IF-OSN \cite{osn}, PSCC-Net \cite{pscc}, CAT-Net 2 \cite{catnet2}, and Hifi \cite{hifi}.
All these methods are with publicly available source codes and re-trained on selected datasets. We automatically choose their hyperparameters as described in the corresponding reference papers or optimally assign better ones.
\begin{figure*}[t]
	\centering 
	\includegraphics[width=0.95\textwidth]{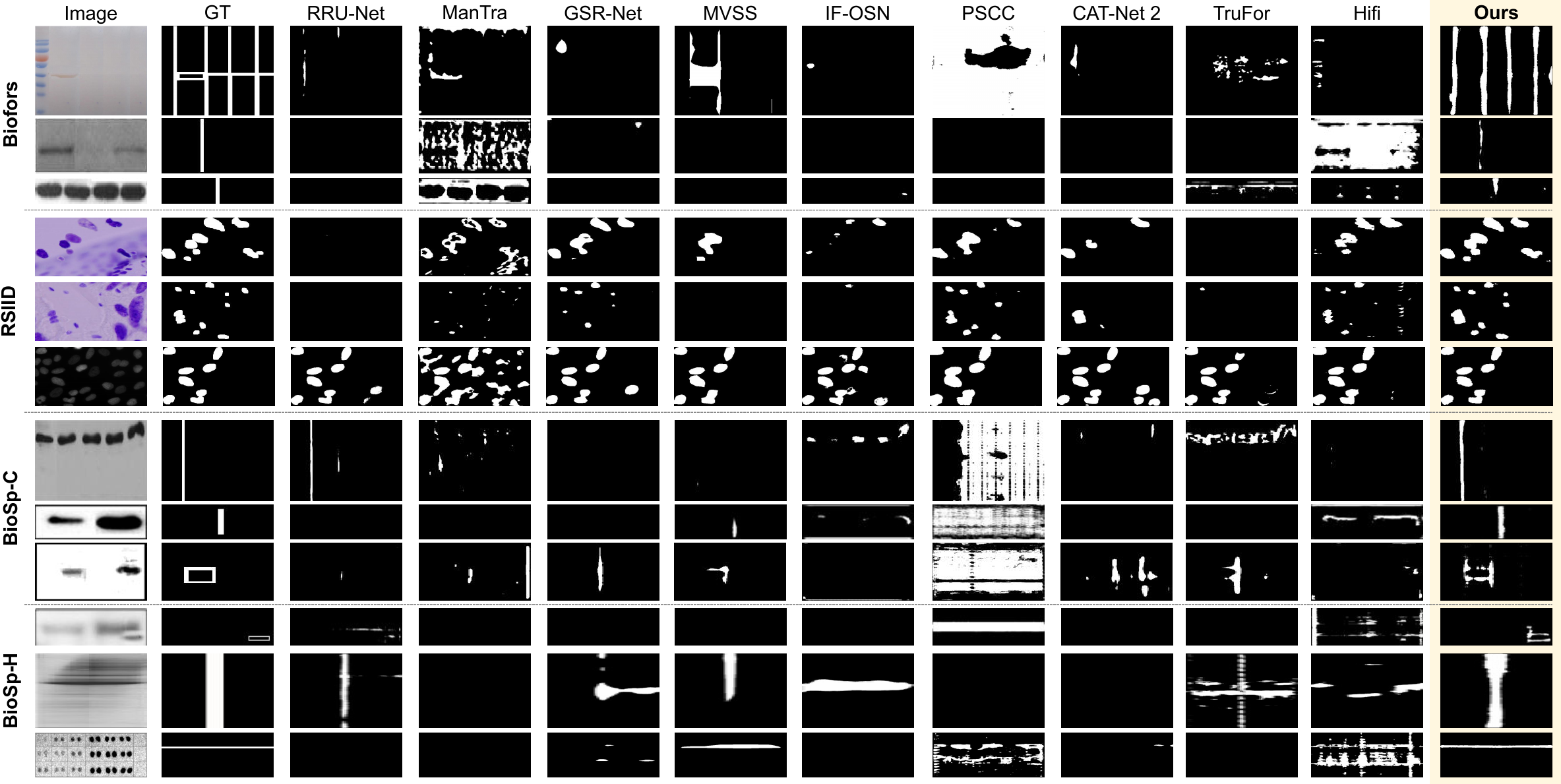}
	\caption{Qualitative results across Biofors, RSIID, SciSp-C, and SciSp-H. From left to right, each column represents images, ground truths (GT), and predictions of competing methods, respectively.}
	\label{fig:vis}
\end{figure*}
\begin{table*}[t]
        \caption{Pixel-level (\ie F1 and MCC) and image-level (\ie AUC and Acc) performance (\%) of scientific image splicing detection. Note that ``ManTra'' represents the results reported in Biofors \cite{biofors}, whereas ``ManTra*'' denotes the results we reproduce.}
	\renewcommand{\arraystretch}{1}
	\label{tab:intra}
	\resizebox*{1.0 \linewidth}{!}{
		\begin{tabular}{l|cccc|cccc|cccc|cccc|cccc}
			\toprule
			\textbf{Datasets $\rightarrow$} & \multicolumn{4}{c|}{\textbf{Biofors}} & \multicolumn{4}{c|}{\textbf{RSIID}} & \multicolumn{4}{c|}{\textbf{SciSp-C}} & \multicolumn{4}{c|}{\textbf{SciSp-H}} & \multicolumn{4}{c}{\textbf{Average}} \\
			\cmidrule(r){1-1} \cmidrule(lr){2-5} \cmidrule(lr){6-9} \cmidrule(lr){10-13} \cmidrule(lr){14-17} \cmidrule(l){18-21}                              
			 \textbf{Methods $\downarrow$} & \textbf{F1} & \textbf{MCC} &  \textbf{AUC} &  \textbf{Acc} & \textbf{F1} & \textbf{MCC} &  \textbf{AUC} &  \textbf{Acc} & \textbf{F1} & \textbf{MCC} &  \textbf{AUC} &  \textbf{Acc} & \textbf{F1} & \textbf{MCC} &  \textbf{AUC} &  \textbf{Acc} & \textbf{F1} & \textbf{MCC} &  \textbf{AUC} &  \textbf{Acc} \\
                \midrule
			% \cmidrule(r){1-1} \cmidrule{2-5} \cmidrule{6-9} \cmidrule{10-13} \cmidrule{14-17} \cmidrule{18-21} 
			RRU-Net\cite{rru} & 14.9 & 14.9 & \underline{79.9} & \textbf{75.5} & \underline{80.6} & \underline{80.4} & 99.1 & 94.0 & 31.9 & 32.7 & \underline{83.0} & \underline{73.3} & \underline{32.4} & \underline{32.8} & 76.1 & 68.0 & 40.0 & 40.2 & \underline{83.4} & \underline{77.7} \\
			ManTra*\cite{mantra} & 12.4 & 8.0 & 55.3 & 50.0 & 48.8 & 49.9 & 98.4 & 84.1 & 20.7 & 21.3 & 73.8 & 63.4 & 0.4 & -0.1 & 54.8 & 59.8 & 20.6 & 20.9 & 69.2 & 64.3 \\
			ManTra\cite{mantra} & 9.0 & 8.0 & - & - & - & - & - & - & - & - & - & - & - & - & - & - & - & - & - & - \\
			GSR-Net\cite{gsr} & 3.1 & 2.4 & 70.3 & 65.7 & 79.1 & 79.0 & 98.8 & 93.3 & 21.7 & 22.2 & 73.4 & 65.8 & 23.6 & 23.4 & 71.1 & 66.8 & 31.9 & 31.9 & 77.2 & 72.9 \\
			MVSS\cite{mvss} & 11.4 & 10.9 & 77.8 & 60.0 & 77.3 & 77.2 & \underline{99.3} & 95.1 & 15.8 & 16.2 & 79.9 & 64.2 & 29.0 & 29.0 & \underline{76.7} & 66.8 & 33.4 & 33.5 & 79.0 & 71.5 \\
			IF-OSN\cite{osn} & 9.6 & 8.7 & 72.8 & 66.7 & 80.1 & 80.0 & 98.9 & 95.1 & 14.3 & 13.5 & 76.7 & 69.8 & 12.5 & 12.0 & 65.3 & 58.0 & 29.1 & 28.8 & 76.9 & 72.4 \\
			PSCC-Net\cite{pscc} & 12.9 & 7.0 & 71.2 & 63.0 & 70.1 & 70.3 & \textbf{99.5} & \textbf{95.4} & 6.5 & 1.5 & 79.4 & 68.0 & 18.3 & 14.7 & 61.9 & 31.2 & 27.0 & 24.9 & 76.0 & 64.4 \\
			CAT-Net 2\cite{catnet2} & 13.8 & 12.4 & 59.8 & 53.7 & 77.4 & 77.1 & 97.9 & 91.2 & 22.9 & 23.0 & 76.8 & 69.8 & 30.4 & 30.3 & 74.6 & 64.0 & 36.1 & 36.1 & 75.8 & 69.7 \\
			TruFor\cite{trufor} & 8.3 & 8.9 & 71.4 & 66.7 & 77.3 & 77.5 & \underline{99.3} & 94.7 & 17.5 & 17.6 & 75.8 & 64.0 & 19.7 & 19.6 & 74.0 & 65.3 & 30.7 & 30.8 & 79.0 & 72.7 \\
			Hifi\cite{hifi} & \underline{21.7} & \underline{20.1} & 66.7 & 61.1 & 77.4 & 77.2 & 98.7 & 84.5 & \underline{35.8} & \underline{37.7} & 77.2 & 72.7 & 26.2 & 26.0 & 66.8 & \textbf{69.0} & \underline{40.3} & \underline{40.7} & 75.9 & 71.8 \\
			\textbf{\model} & \textbf{30.3} & \textbf{30.4} & \textbf{87.8} & \underline{75.0} & \textbf{81.4} & \textbf{81.4} & \underline{99.3} & \underline{94.9} & \textbf{38.5} & \textbf{39.3} & \textbf{85.9} & \textbf{75.0} & \textbf{38.9} & \textbf{38.9} & \textbf{78.0} & \underline{68.7} & \textbf{47.3} & \textbf{47.5} & \textbf{84.6} & \textbf{78.4} \\
			\bottomrule[1.3pt]
		\end{tabular}
	}
\end{table*}

In \tabref{tab:intra}, we show both pixel-level and image-level detection results. Our method outperforms all SoTAs in terms of average metric values across all datasets, achieving the highest scores in both pixel-level metrics for each dataset. Besides, half of the image-level metric values of \model~rank second. As we can see in \figref{fig:vis}, our method has lower false alarms and is more capable of accurately localizing subtle splicing traces. This indicates the effectiveness of our strategy of using uncertainty information to refine predictions.
Besides, images in RSIID have fewer disruptive factors, such as noises and artifacts, which are less likely to mislead the detection method into making incorrect predictions.  The superior performance of our method on SciSp-H further demonstrates that the effective use of uncertainty can mitigate the effect of disruptive factors, thus enhancing robustness.

\begin{figure*}[t]
	\centering 
	\includegraphics[width=1.0\textwidth]{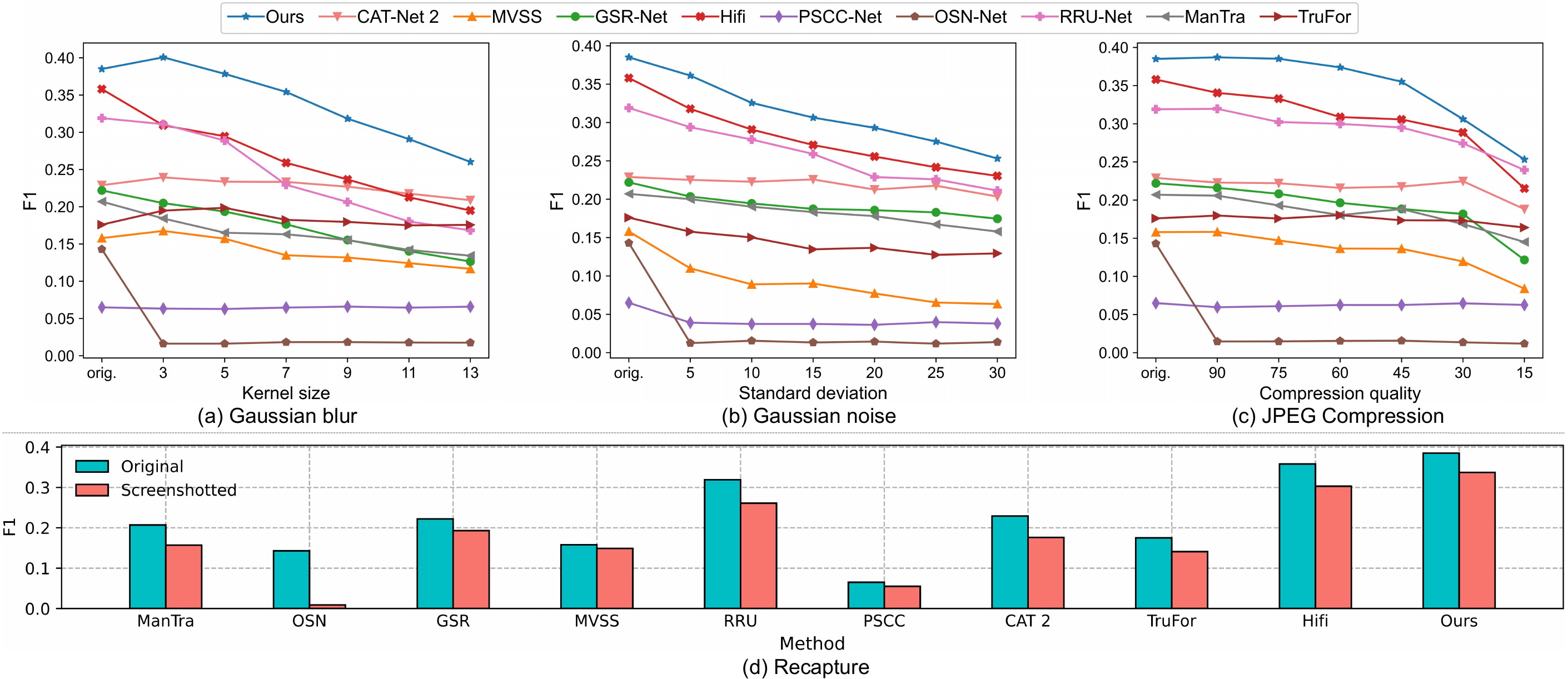}
	\caption{Robustness analysis results on SciSp-C against four different degradations, \ie (a) Gaussian blur, (b) Gaussian noise, (c) JPEG compression, and (d) recaptured by Windows Snipping Tool.} 
	\label{fig:degrade} 
\end{figure*}
\begin{figure}[t]
	\centering 
	\includegraphics[width=0.5\textwidth]{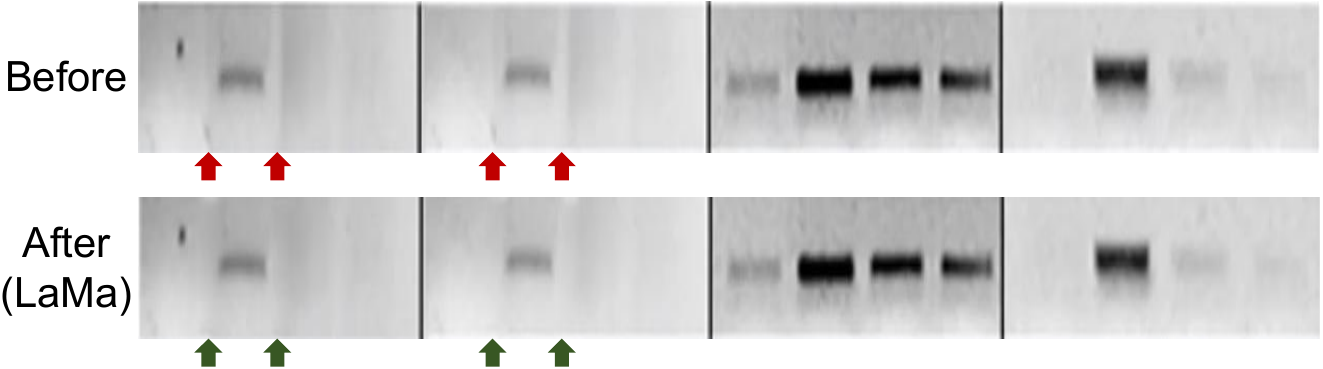}
	\caption{Example of using the advanced AI-based inpainting scheme, \textit{i.e.} LaMa \cite{lama}, to decrease splicing traces in the first two panels. The arrows indicate the splicing traces. The source image is from Biofors.}
	\label{fig:lama}
\end{figure}
\begin{table*}[t]
	\caption{Robustness comparisons against four inpainting approaches (\ie~LaMa, Stable diffusion v2, Navier-stokes, and Telea) on SciSp-C. }
	\center
	\renewcommand{\arraystretch}{1}
	\label{tab:inpainting}
	\resizebox*{1.0 \linewidth}{!}{
        \begin{tabular}{@{}l|cccc|cccc|cccc|cccc|cccc@{}}
    		\toprule
    		 \textbf{Inpainting} $\rightarrow$ & \multicolumn{4}{c|}{\textbf{LaMa\cite{lama}}} & \multicolumn{4}{c|}{\textbf{Stable diffusion v2\cite{diffusion}}} & \multicolumn{4}{c|}{\textbf{Navier-stokes\cite{navier}}} & \multicolumn{4}{c|}{\textbf{Telea\cite{telea}}} & \multicolumn{4}{c}{\textbf{Average}} \\ 
    		\cmidrule(r){1-1}\cmidrule(lr){2-5} \cmidrule(lr){6-9} \cmidrule(lr){10-13} \cmidrule(lr){14-17} \cmidrule(l){18-21}                              
    		\textbf{Method} $\downarrow$ & \textbf{F1} & \textbf{MCC} & \textbf{AUC} & \textbf{Acc} & \textbf{F1} & \textbf{MCC} & \textbf{AUC} & \textbf{Acc} & \textbf{F1} & \textbf{MCC} & \textbf{AUC} & \textbf{Acc} & \textbf{F1} & \textbf{MCC} & \textbf{AUC} & \textbf{Acc} & \textbf{F1} & \textbf{MCC} & \textbf{AUC} & \textbf{Acc} \\ 
    		\cmidrule(l){1-1} \cmidrule{2-5} \cmidrule{6-9} \cmidrule{10-13} \cmidrule{14-17} \cmidrule{18-21}
    		RRU-Net\cite{rru} & 15.6 & 15.8 & 73.9 & \underline{66.9} & 31.9 & 32.7 & \underline{83.0} & \underline{73.3} & 25.2 & 25.6 & 79.9 & \underline{70.4} & 20.4 & 21.0 & \underline{77.1} & 68.0 & 23.3 & 23.8 & \underline{78.5} & 69.6 \\
    		ManTra\cite{rru} & 8.2 & 8.2 & 68.2 & 62.8 & 20.7 & 21.3 & 73.8 & 63.4 & 15.2 & 15.4 & 70.7 & 61.6 & 11.5 & 11.5 & 69.4 & 62.2 & 13.9 & 14.1 & 70.5 & 62.5 \\
    		GSR-Net\cite{gsr} & 12.7 & 13.2 & 68.0 & 66.3 & 21.7 & 22.2 & 73.4 & 68.6 & 15.1 & 15.7 & 69.5 & 68.0 & 13.6 & 14.4 & 69.0 & 67.4 & 15.7 & 16.4 & 70.0 & 67.6 \\
    		MVSS\cite{mvss} & 12.1 & 12.2 & 75.0 & 62.8 & 15.8 & 16.2 & 79.9 & 69.8 & 13.4 & 14.4 & \underline{80.1} & 68.6 & 12.4 & 13.0 & 78.4 & 66.9 & 13.4 & 14.0 & 78.3 & 67.0 \\
    		IF-OSN\cite{osn} & 1.0 & 1.3 & 68.2 & 62.8 & 1.5 & 1.9 & 68.2 & 62.8 & 1.0 & 1.2 & 68.0 & 62.8 & 1.0 & 1.2 & 67.8 & 62.8 & 1.1 & 1.4 & 68.1 & 62.8 \\
    		PSCC-Net\cite{pscc} & 5.8 & 2.8 & 58.2 & 50.0 & 5.9 & 3.1 & 58.6 & 50.0 & 6.1 & 3.4 & 58.6 & 50.0 & 5.9 & 3.3 & 58.2 & 50.0 & 5.9 & 3.1 & 58.4 & 50.0 \\
    		CAT-Net 2\cite{catnet2} & 20.9 & 20.8 & \underline{75.2} & 69.2 & 22.9 & 23.0 & 76.8 & 69.8 & 21.3 & 21.4 & 75.6 & \underline{70.9} & 20.4 & 20.4 & 75.7 & 70.9 & 21.4 & 21.4 & 75.8 & \underline{70.2} \\
    		TruFor\cite{trufor} & 14.6 & 14.8 & 70.2 & 67.4 & 17.5 & 17.6 & 71.7 & 68.0 & 13.1 & 13.3 & 69.8 & 68.0 & 13.1 & 13.1 & 69.1 & 67.4 & 14.6 & 14.7 & 70.2 & 67.7 \\
    		Hifi\cite{hifi} & \underline{23.5} & \underline{24.1} & 73.0 & 50.0 & \underline{35.8} & \underline{37.7} & 73.8 & 50.0 & \underline{29.8} & \underline{32.0} & 73.8 & 50.0 & \underline{28.1} & \underline{29.9} & 73.4 & 50.0 & \underline{29.3} & \underline{30.9} & 73.5 & 50.0 \\
    		\textbf{\model} & \textbf{28.6} & \textbf{28.9} & \textbf{82.1} & \textbf{73.8} & \textbf{40.1} & \textbf{40.6} & \textbf{85.7} & \textbf{75.0} & \textbf{36.0} & \textbf{36.7} & \textbf{84.8} & \textbf{74.4} & \textbf{32.5} & \textbf{32.9} & \textbf{84.2} & \textbf{72.1} & \textbf{34.3} & \textbf{34.8} & \textbf{84.2} & \textbf{73.8} \\
    		\bottomrule[1.3pt]
	\end{tabular}
		}
\end{table*}
\begin{table*}[t]
	\caption{Pixel-level (\ie F1 and MCC) and image-level (\ie AUC and Acc) \textbf{cross-testing results} (\%). ``B'', ``C'', and ``H'' represent Biofors, SciSp-C, and SciSp-H, respectively. ``X + Y $\rightarrow$ Z'' denotes that the union of X and Y are used for training and Z is adopted as the test set. The first and second rankings are respectively shown in bold and underlined.}
	\center
	\renewcommand{\arraystretch}{1.0}
	% \tabSpace
	\label{tab:cross}
	\resizebox*{1.0 \linewidth}{!}{
		\begin{tabular}{l|cccc|cccc|cccc|cccc}
			\toprule
			\textbf{Dataset} $\rightarrow$ & \multicolumn{4}{c|}{\textbf{B + C $\rightarrow$ H}} & \multicolumn{4}{c|}{\textbf{C + H $\rightarrow$ B}} & \multicolumn{4}{c|}{\textbf{H + B $\rightarrow$ C}} & \multicolumn{4}{c}{\textbf{Average}} \\
			\cmidrule(r){1-1} \cmidrule(lr){2-5} \cmidrule(lr){6-9} \cmidrule(lr){10-13} \cmidrule(l){14-17}
			 \textbf{Method} $\downarrow$ & \textbf{F1} & \textbf{MCC} &  \textbf{AUC} &  \textbf{Acc} & \textbf{F1} & \textbf{MCC} &  \textbf{AUC} &  \textbf{Acc} & \textbf{F1} & \textbf{MCC} &  \textbf{AUC} &  \textbf{Acc} & \textbf{F1} & \textbf{MCC} &  \textbf{AUC} &  \textbf{Acc} \\
                \midrule
			RRU-Net\cite{rru} & 13.1 & 12.6 & 56.2 & 53.8 & 22.1 & 23.4 & \underline{70.8} & \underline{69.4} & 21.5 & 22.6 & \underline{76.0} & 66.9 & 18.9 & 19.6 & 67.7 & 63.4 \\
			ManTra\cite{mantra} & 9.0 & 2.5 & 49.3 & 31.2 & 0.2 & 0.1 & 50.1 & 52.8 & 10.2 & 8.2 & 47.6 & 50.0 & 6.5 & 3.6 & 49.0 & 44.7 \\
			GSR-Net\cite{gsr} & 3.7 & 3.4 & 52.8 & 51.8 & 15.2 & 15.4 & 67.7 & 63.9 & 1.8 & 1.5 & 60.9 & 56.4 & 6.9 & 6.8 & 60.4 & 57.4 \\
			MVSS\cite{mvss} & 11.2 & 11.0 & \textbf{60.5} & \textbf{60.7} & 20.8 & 21.3 & 75.3 & \underline{69.4} & 13.7 & 14.3 & 79.8 & \underline{71.5} & 15.3 & 15.5 & \underline{71.9} & \underline{67.2} \\
			IF-OSN\cite{osn} & 9.2 & 8.5 & 51.9 & 47.0 & 8.1 & 7.8 & 52.5 & 60.2 & 4.9 & 5.2 & 63.9 & 59.9 & 7.4 & 7.1 & 56.1 & 55.7 \\
			PSCC-Net\cite{pscc} & 11.8 & 8.4 & 54.0 & 31.2 & 12.5 & 5.0 & 60.4 & 50.0 & \textbf{28.0} & \textbf{31.4} & 58.8 & 50.0 & 17.4 & 14.9 & 57.7 & 43.7 \\
			CAT-Net 2\cite{catnet2} & 10.1 & 8.3 & 55.1 & 43.2 & 11.7 & 11.8 & 64.0 & 59.3 & 12.8 & 12.9 & 65.3 & 61.6 & 11.5 & 11.0 & 61.5 & 54.7 \\
			TruFor\cite{trufor} & 6.2 & -0.4 & 51.0 & 33.2 & 11.0 & 7.9 & 47.6 & 49.1 & 3.7 & 1.7 & 51.9 & 50.6 & 7.0 & 3.1 & 50.2 & 44.3 \\
			Hifi\cite{hifi} & \textbf{17.1} & 13.2 & 48.4 & 31.2 & \textbf{27.7} & \underline{26.3} & 61.1 & 50.0 & 18.7 & 22.3 & 59.2 & 50.0 & \underline{21.2} & \underline{20.6} & 56.2 & 43.7 \\
			\textbf{\model} & \underline{15.1} & \textbf{14.6} & \underline{59.4} & \underline{55.8} & \underline{25.8} & \textbf{26.8} & \textbf{82.9} & \textbf{77.8} & \underline{23.4} & \underline{24.0} & \textbf{80.9} & \textbf{75.0} & \textbf{21.4} & \textbf{21.8} & \textbf{74.4} & \textbf{69.5} \\
			\bottomrule[1.3pt]
		\end{tabular}
	}
\end{table*}

\subsection*{Robustness against post-processing}
In this section, we conduct experiments to verify the robustness of our method in handling different forms of post-processing, \ie degradation and inpainting. For a fair verification, we only utilized post-processing approaches on the testing set.

\paragraph{Capability to resist degradation}
In real-world scenarios, scientific images are always degraded by multiple approaches. Here, we simulate these conditions with various degrees of commonly used degradations. 
For each degradation approach, we apply various parameter values, including the kernel size of Gaussian blur (from 3 to 13), the quality of JPEG compression (from 15\% to 90\%), and the standard deviation of Gaussian noise (from 5 to 30), to make a comprehensive verification. In addition, we evaluate performance on images recaptured by the Windows Snipping Tool on a Windows 10 computer with a screen resolution of  1920$\times$1080.
As observed from the \figref{fig:degrade}, our method maintains the best performance under various degradation attacks. Even under the most severe degradation, our performance exceeds some other methods under no degradation. 

\paragraph{Capability to resist inpainting}
Inpainting is widely used for hidden visual clues of image manipulation \cite{csvt-inpaint}. To validate the resistance of models against inpainting attacks, we employed two state-of-the-art generative algorithms, \ie LaMa \cite{lama} and Stable diffusion v2 \cite{diffusion} and two commonly used traditional algorithms, Navier-stokes \cite{navier} and Telea \cite{telea} to hide the spliced regions. To the best of our knowledge, this kind of experiment has not been performed before in the field of image manipulation detection. As shown in \tabref{tab:inpainting}, we achieved the best performance when dealing with every involved inpainting method.\\

\subsection*{Generalisability against domain shifts}
In real-world applications, methods are challenged by domain shifts and unseen splicing approaches. To further validate the generalisability of our model, we conduct this cross-dataset testing. It can be seen in \tabref{tab:cross} that our URN achieves the best performance in average metric values on all protocols. \model~reaches the highest scores on half of the metrics and ranks second on the rest. This indicates that our method can resist domain shifts and unseen attacks using reliable information from confident regions. 

\section*{Discussion}
It can be seen from Fig.~\ref{fig:degrade} and Tables~\ref{tab:intra}-\ref{tab:cross}, outstanding detecting performance, robustness, and generalisability of URN. 
This can be attributed to our effective utilization of uncertainty information. 
Even if decreases the confidence of predictions in some areas, our stage-2 network integrated with proposed UEMA and UGGC can refine them to yield inspiring results.
In the following sub-sections, we provide the ablation study, limitations, and social impact of this work, respectively.
\begin{figure*}[t]
	\centering 
	\includegraphics[width=1.0\textwidth]{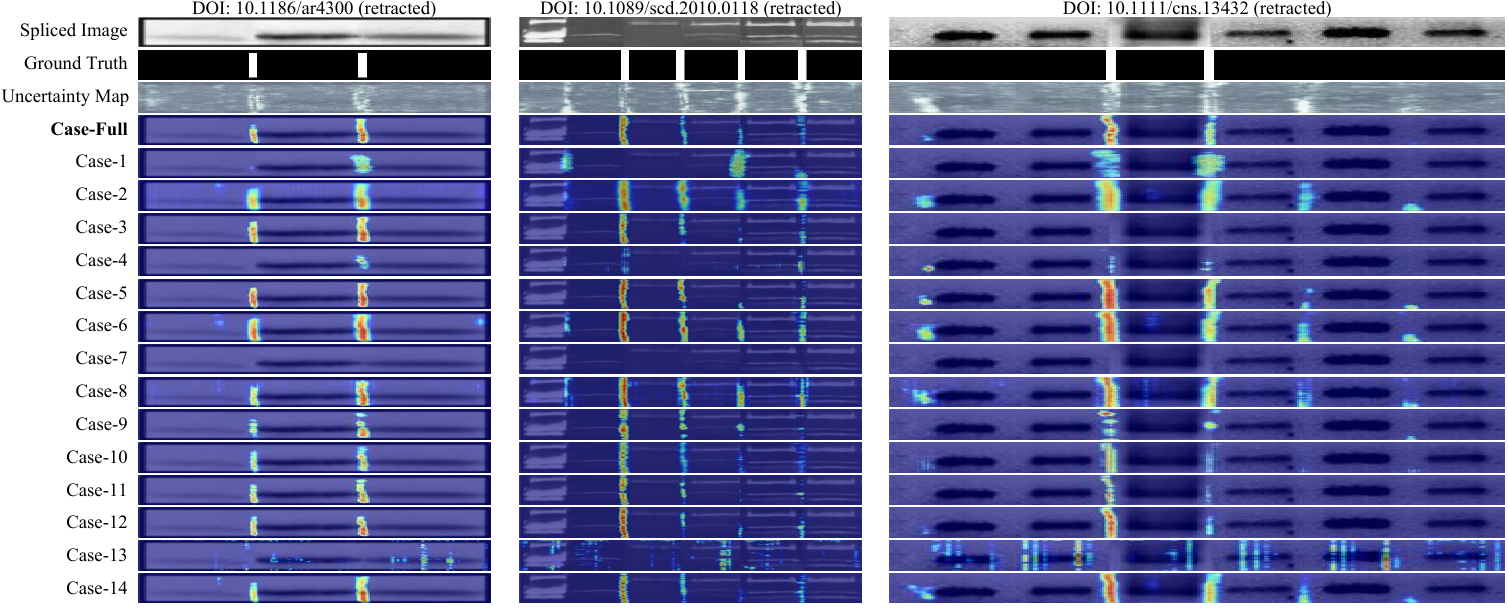}
	\caption{Feature visualization of ablation study. From top to bottom, each row respectively represents the spliced images, ground truths, uncertainty maps, and feature heatmaps of the corresponding case.}
	\label{fig:abla}
\end{figure*}
\begin{table}[t]
        \centering
	\renewcommand{\arraystretch}{1.0}
        \setlength{\tabcolsep}{15pt}
	\caption{Ablation results (\%) on SciSp-C.}
	% \tabSpace
	\label{tab:abla}
	\resizebox*{1.0 \linewidth}{!}{
		\begin{tabular}{ccccccccc}
                \toprule[0.1pt]
			\toprule
%			\begin{tabular}[c]{@{}c@{}}\textbf{Setup}\\ \textbf{Name}\end{tabular}
			\multirow{3}{*}{\textbf{Case}} & \multirow{3}{*}{\textbf{Refinement}} & \multicolumn{5}{c}{\textbf{Stage 2 Refinement Modules}} & 
			\multicolumn{2}{c}{\multirow{2}{*}{\begin{tabular}[c]{@{}c@{}}\textbf{Metric}\\ \textbf{Values}\end{tabular}}} \\ 
			\cmidrule(lr){3-7} 
			&  & \multicolumn{4}{c}{\textbf{Encoding}} & \textbf{Decoding} & \multicolumn{2}{c}{} \\ 
			\cmidrule(lr){3-6} \cmidrule(lr){7-7} \cmidrule(lr){8-9}
			&  & \textbf{Module} & \textbf{Local} & \textbf{Directed} & \textbf{Weighted} & \textbf{Module} & \textbf{F1} & \textbf{MCC} \\ 
                \midrule[0.1pt]
                \midrule
			% \cmidrule(lr){1-1} \cmidrule(lr){2-2} \cmidrule{3-6} \cmidrule(lr){7-7}  \cmidrule{8-9}
			1 & \tiny\XSolid & - & - & - & - & - & 30.4 & 31.9 \\ \midrule
			2 & \tiny\Checkmark & - & - & - & - & UEMA & 32.0 & 34.5 \\
			3 & \tiny\Checkmark & UGGC (kNN\cite{vig}) & - & \tiny\Checkmark & \tiny\XSolid & UEMA & 30.0 & 31.0 \\
			4 & \tiny\Checkmark & UGGC & \tiny\Checkmark & \tiny\XSolid & \tiny\XSolid & UEMA & 21.6 & 22.6 \\
			5 & \tiny\Checkmark & UGGC & \tiny\Checkmark & \tiny\Checkmark & \tiny\XSolid & UEMA & 33.8 & 35.3 \\
			6 & \tiny\Checkmark & Self-Attention & \tiny\XSolid & \tiny\XSolid & \tiny\XSolid & UEMA & 32.5 & 34.9 \\
			7 & \tiny\Checkmark & UGGC & \tiny\XSolid & \tiny\Checkmark & \tiny\XSolid & UEMA & 33.2 & 34.3 \\
			8 & \tiny\Checkmark & UGGC & \tiny\XSolid & \tiny\Checkmark & \tiny\Checkmark & UEMA & 35.8 & 37.0 \\ \midrule
			9 & \tiny\Checkmark & UGGC & \tiny\Checkmark & \tiny\Checkmark & \tiny\Checkmark & - & 32.5 & 33.6 \\
			10 & \tiny\Checkmark & UGGC & \tiny\Checkmark & \tiny\Checkmark & \tiny\Checkmark & Self-Attention & 35.0 & 35.5 \\
			11 & \tiny\Checkmark & UGGC & \tiny\Checkmark & \tiny\Checkmark & \tiny\Checkmark & Cross-Attention & 29.0 & 30.4 \\
			12 & \tiny\Checkmark & UGGC & \tiny\Checkmark & \tiny\Checkmark & \tiny\Checkmark & Depth-wise Convolution & 34.6 & 35.6 \\
			13 & \tiny\Checkmark & UGGC+UEMA & \tiny\Checkmark & \tiny\Checkmark & \tiny\Checkmark & UEMA & 36.9 & 37.3 \\
			14 & \tiny\Checkmark & UGGC+UEMA & \tiny\Checkmark & \tiny\Checkmark & \tiny\Checkmark & - & 34.6 & 35.5 \\ \midrule
			\textbf{Full} & \textbf{\tiny\Checkmark} & \textbf{UGGC} & \textbf{\tiny\Checkmark} & \textbf{\tiny\Checkmark} & \textbf{\tiny\Checkmark} & \textbf{UEMA} & \textbf{38.5} & \textbf{39.3}\\ 
			\bottomrule
		\end{tabular}
	}
\end{table}
\subsection*{Ablation Study on URN}
To comprehensively validate the individual effectiveness of components of \model, we not only compare the performance of our model before and after the inclusion of key modules, but we also conduct comparisons between multiple variants of important sub-modules. 
In \tabref{tab:abla}, we present the specific configurations and performance of different variants by replacing or removing key components.
To further confirm the regions each case tends to focus on, we show feature visualization results of all cases using Grad-CAM \cite{gradcam} in \figref{fig:abla}.

\paragraph{Effectiveness of the two-stage framework.}
From the row UM in \figref{fig:abla}, it can be seen that the uncertainty maps we estimated in the stage-1 network effectively identify regions prone to erroneous prediction due to disruptive factors. As shown in \tabref{tab:abla}, the metric values of Case 1 are lower than most of the other cases. Additionally, from the 4th row in \figref{fig:abla}, we can see that Case 1 loses focus on some spliced regions and incorrectly attends to pristine areas. 
This indicates that appropriately leveraging uncertainty information can help improve performance.

\paragraph{Effectiveness of the UGGC.}
As shown in \tabref{tab:abla}, we observe that adding UGGC modules (compare Case 2 with Full) brings a 6.5\% and 4.8\% increase in F1 and MCC, respectively. Also, it helps to reduce the erroneous attention to non-spliced regions with high uncertainty (see \figref{fig:abla}). 
This demonstrates that UGGC can assist \model~in learning more robust features.
Additionally, in Cases 3-8, we investigate the impact of different edge connection strategies for the construction of $\mathcal G^k$ in UGGC, including the scope of edge connection, and whether the edges are directed and weighted. As shown in \tabref{tab:abla}, regardless of the type of involved edge connection strategy used, all resulted in a decrease in performance compared to the proposed strategy.
This demonstrates that restricting the information flow from confident nodes to local uncertain ones can help weak-predicted regions capture local inconsistency from nearby areas, thereby enhancing robustness.

\paragraph{Effectiveness of the UEMA.}
As shown in \tabref{tab:abla} and \figref{fig:abla}, we explored the effectiveness of UEMA by constructing Cases 9-12. Whether UEMA is removed or replaced with self-attention, cross-attention, or depth-wise convolution, none can achieve better performance. It should be noted that the metric values of Case 11 drop, indicating that a semantic gap exists between estimated uncertainty maps and features in deep layers. Directly calculating their correlation would mislead the decoding process. 
This also demonstrates that our strategy of enhancing feature maps with uncertainty maps can help UEMA focus on uncertain regions and distinguish between spliced and pristine areas.
To further explore the effectiveness of UEMA in different positions of our \model, we give a comparison among Cases 13, 14, and Full in \tabref{tab:abla} and \figref{fig:abla}. We testify two variants: Case 13 with UEMA in both the decoding and encoding stages, and Case 14 with UEMA only in the encoding stage. We observe that the performance of Case 13 is inferior to Case Full, suggesting that the unrestricted global information transition in UEMA interferes with the robust features refined by UGGC during encoding. 
Moreover, the performance of Case 14 is even worse than Case 13, implying that additional attention to uncertain regions is effective during the decoding stage.
% \noindent The above experimental results demonstrate the high robustness of our method against various post-processing schemes. This can be attributed to our effective utilization of uncertainty information. Even if decreases the confidence of predictions in some areas, our stage-2 network integrated with proposed UEMA and UGGC can refine them to yield inspiring results.

% However, the cross-dataset results are not as inspiring as the intra-dataset ones, which motivates us to explore techniques to improve the domain generalisability. As for future research, we aim to enable existing methods can be easily adapted to other types of biomedical images or even images from other fields (\eg materials science, chemistry, and mechanical engineering). 

\subsection*{Limitations of the study}
As shown in Table \ref{tab:cross}, the cross-dataset results are not as inspiring as the intra-dataset ones (see Table \ref{tab:intra}), which motivates us to explore techniques to improve the domain generalisability. 
As for future research, we aim to further improve the domain generalization ability of our method, enabling it can be easily adapted to other types of scientific images or even images from other fields (\eg materials science, chemistry, and mechanical engineering). 
Additionally, there is a significant imbalance between the number of pristine and spliced images. How to effectively leverage the vast number of pristine images is also an interesting direction for further investigation.

\subsection*{Social impact}
Our proposed method and dataset yield both beneficial and adverse societal impacts, with the former believed to surpass the latter. The primary benefit lies in the ability of our method to expose splicing traces in scientific images, preventing misleading academic peers and upholding scholarly integrity. However, the present accuracy of our method remains inadequate, necessitating expert review for result validation in real-world applications. Besides, false alarms given by our method may be exploited by malicious individuals to unjustly accuse others' publications. Last but not least, the existence of our method may compel some researchers to meticulously manipulate images via more advanced techniques.

\section*{Experimental procedures}
\begin{figure*}[t]
	\centering 
	\includegraphics[width=1.0\textwidth]{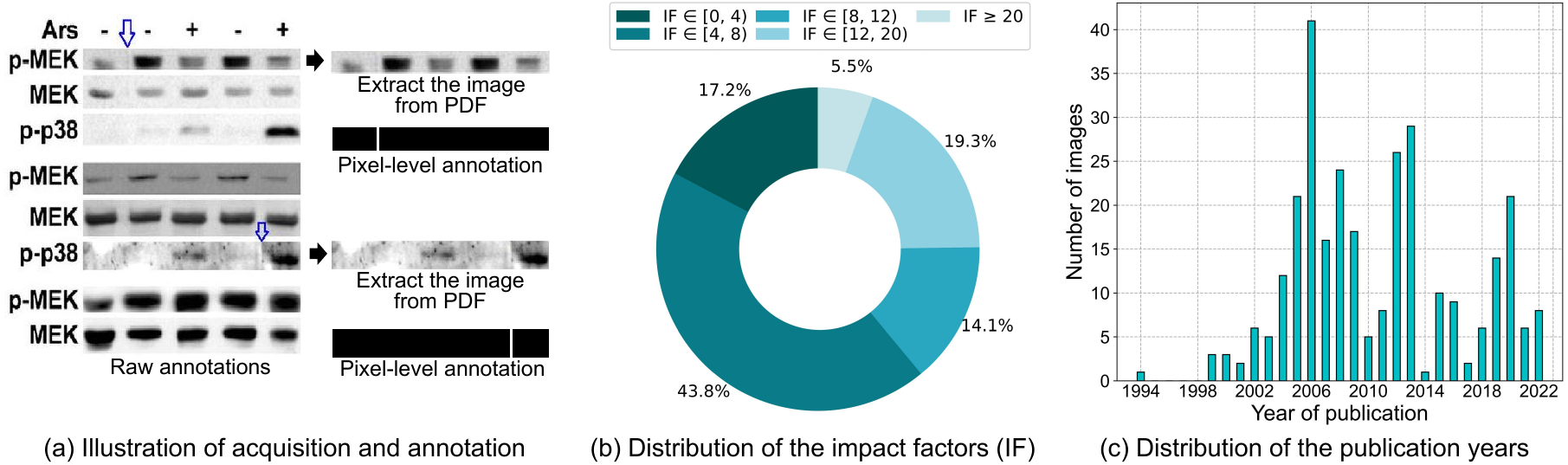}
	\caption{Details of SciSp-C. In the left part of (a), the academic peer uses blue arrows to expose splicing traces. We make pixel-level annotations according to these raw annotations. (b) Distribution of the impact factors of journals to which the images in SciSp-C belong. (c)Distribution of the publication years of papers to which the images in SciSp-C belong.} 
	\label{fig:biosp-c} 
\end{figure*}
\begin{figure*}[t]
	\centering 
	\includegraphics[width=0.95\textwidth]{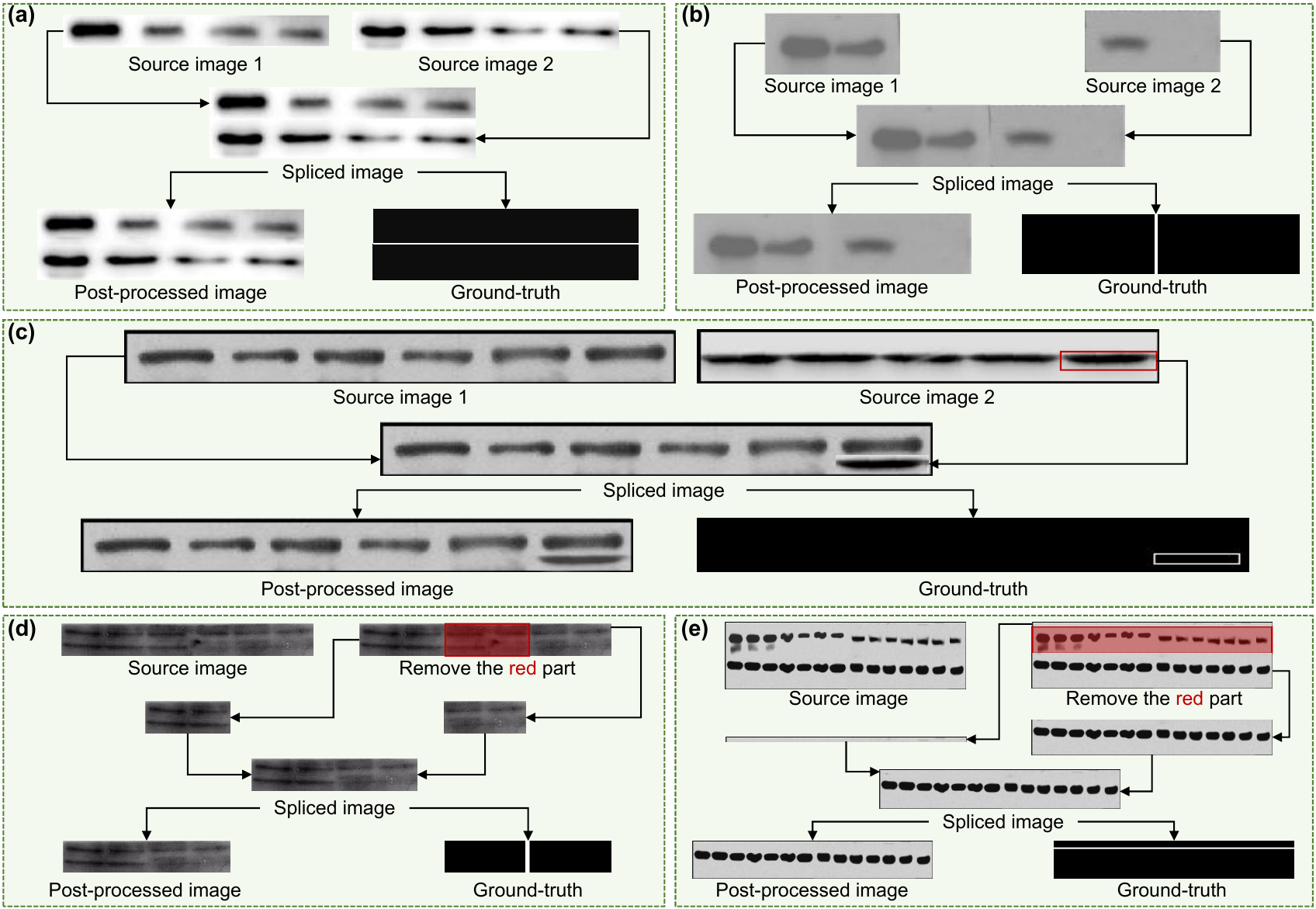}
	\caption{Illustration of five splicing approaches, \ie (a) vertical splicing, (b) horizontal splicing, (c) free splicing, (d) vertical removal and splicing, and (e) horizontal removal and splicing, used to construct SciSp-H. } %最终文档中希望显示的图片标题
	\label{fig:biosp-h} %用于文内引用的标签
\end{figure*}
\subsection*{Datasets}
High-quality datasets are essential to advance the field of scientific splicing detection. 
Most scientific images have not been proven to be improperly manipulated, even if the image is indeed forged, which makes it difficult for data collection.
Here, we construct the two sub-datasets: (1) we collect images published in various journals according to the public comments from Pubpeer, and (2) we manually forge images with multiple splicing approaches and perform post-processing to reduce visible traces.
\paragraph{SciSp-C: Collected Images.} 
We collected 110 public comments posted on PubPeer, pointing out that 290 images show signs of splicing. 
Among these images, 42 are from retracted papers, 102 are from corrected ones, and 27 have publicly acknowledged splicing. 
The remaining images are deemed as spliced because the authors cannot provide solid evidence to rebut the concerns, such as being unable to present the original images.
All collected images are with raw annotations provided by academic peers, suggesting that the images are highly likely to be spliced.

The annotation process is shown in \figref{fig:biosp-c}(a), where the raw annotation is collected from Pubpeer\footnote{\url{https://pubpeer.com/publications/DC5E463CDD7A38F6A22A3D5325B800}}. First, we find raw annotations related to spliced scientific images from public comments provided by experts. To prevent image degradation caused by screenshots, we download the corresponding involved papers (PDF version) and use Kingsoft WPS\footnote{https://www.wps.com/} to extract spliced images. Finally, according to the raw annotations, we use Labelme\footnote{https://github.com/wkentaro/labelme} to annotate the pixel-level binary masks of spliced regions. Given that neither we nor the experts know which parts of an image originate from other images, we are unable to annotate all the external spliced areas as in natural image datasets. Following the annotation manner of Biofors, we only mark the junctions of spliced regions. In contrast, images from RSIID are all automatically generated images, with complete external regions annotated in their ground-truths. This results in a different form of ground-truth binary masks compared to Biofos and our SciSp.

As shown in Figs. \ref{fig:biosp-c}(b)-(c), SciSp-C has diverse sources of journals and publication years. It also includes journals of all tiers of influence. Different journals, or the same journal across different years, are likely to use different degradation methods. Therefore, expanding the diversity of sources is important for improving the generalisability of detection models. Most of the images in SciSp-C have been retracted by the publisher or corrected by the authors. For the remaining images, reviewers have provided substantial evidence of splicing traces.

\paragraph{SciSp-H: Handcrafted Images.}
Given that it is difficult to build a large dataset based on the limited number of publicly questioned images, RSIID automatically splices 880 images according to its pre-set rules. However, the spliced region of images in RSIID is easily detectable because the images lacked refined post-processing. Consequently, we manually construct the SciSp-H with 1,000 spliced images to make the dataset more challenging.

To diversify the splicing approaches, we design five different approaches to generate images, \ie (1) vertical splicing: splice two images in the vertical direction, (2) horizontal splicing: splice two images in the horizontal direction, (3) free splicing: select a part from one image and then splice it onto another image, (4) vertical removal and splicing: remove a vertical part of an image and then horizontally splice the remaining two parts together, and (5) horizontal removal and splicing: remove a horizontal part of an image and then vertically splice the remaining two parts together. The detailed illustration of these five splicing approaches is shown in \figref{fig:biosp-h}. For each approach, we create 200 images, all of which are meticulously post-processed to minimize noticeable visual traces. 
Since Adobe Photoshop has been widely used for image manipulation, we adopt a version of CC 2019 to manually generate spliced images. During the process of image splicing and post-processing, the involved tools of Photoshop include \textit{content-aware filling}, \textit{spot healing brush}, \textit{brush}, \textit{cutout}, \textit{blur}, \textit{sharpen}, and \textit{desaturate}. We note that all source images in SciSp-H come from images identified as pristine in Biofors.

\paragraph*{Datasets for evaluation.} 
In addition to our two sub-datasets, \ie SciSp-C and SciSp-H, we also conduct experiments on two publicly available scientific datasets, \ie Biofors (cut/sharp-transition part) and RSIID (splicing part), that are applicable for splicing detection. 
To prevent the interference of false negatives, we randomly sample pristine images from Biofors for SciSp. We ensure that there is no overlap in the selection of pristine images among Biofors, SciSp-C, and SciSp-H. To avoid biases due to imbalanced data, we sample pristine images to ensure their quantity is equal to spliced ones. Note that we adopt a training-testing ratio of 7:3 for each dataset. 

\subsection*{Overall framework of URN}
To capture local inconsistencies caused by splicing, we adopt CNN as the backbone due to its strong ability of local detail extraction. As illustrated in \figref{fig:overall}, our \model~consists of two stages to estimate pixel-level uncertainty maps and perform prediction refinement, respectively. Given a RGB image $\boldsymbol x_i \in \mathbb{R}^{H\times W \times 3}$ as input, the proposed stage-1 network forwards the encoder-decoder to form the binary coarse mask $\boldsymbol y_m \in \{0,1\}^{H\times W}$ and uncertainty map $\boldsymbol y_u \in \mathbb{R}^{H\times W}$. Then the stage-2 network takes $\boldsymbol y_m$ and $\boldsymbol x_i$ as joint inputs, and $\boldsymbol y_u$ is integrated into all encoder and decoder blocks, ultimately predicting a fine mask $\boldsymbol y_v \in \{0,1\}^{H\times W}$. Each encoder unit reduces the height and width of its input to half, while each decoder unit doubles them.

To distinguish spliced and pristine regions, developing neural networks (NNs) is an intuitive solution. However, traditional NNs lack the ability to accurately estimate the uncertainty of detection results. Besides, when dealing with images with disruptive factors or out-of-distribution data, they tend to provide wrong predictions with overconfidence. Therefore, we integrate Monte Carlo Dropout (MCD) layers to identify weakly predicted pixels with high uncertainty. Specifically, following Bayesian SegNet \cite{bayesian}, we integrate an MCD layer after each encoder and decoder unit. 
During inference, the dropouts are still kept active, allowing the sampling of multiple predictions $\mathcal{S} =\{\boldsymbol y^1_s, \boldsymbol y^2_s, \boldsymbol y^3_s,\cdots, \boldsymbol y^{n_s}_s \}$ where $n_s$ is the total number of the samples.
We take the average value of the samples in $\mathcal{S}$ as the coarse mask $\boldsymbol y_m = \sum_{i=1}^{n}\boldsymbol y^i_s/n_s$ and use their normalized variance to represent the uncertainty map $\boldsymbol y_u = sigmoid(\sqrt{\sum_{i=1}^{n_s}(\boldsymbol y^i_s - \boldsymbol y_m)^2/(n_s-1)})$.

\subsection*{Uncertainty-Guided Refinement}
Uncertainty can accurately reflect confidence in the prediction results of deep learning-based methods and has been widely applied in computer vision tasks such as semantic segmentation \cite{aaai_uc_seg}.
The success of uncertainty learning in other fields has also promoted its application for image forensics tasks, \ie JPEG artifacts \cite{jpeg-uc}, resample artifacts \cite{resampling-uc}, and evaluating the confidence of natural image forgery detection results \cite{trufor}. 

Motivated by their works, we introduce uncertainty estimation for robust splicing feature representation learning. Unlike the aforementioned uncertainty-integrated methods, we make full use of uncertainty to recognize and refine uncertain predicted parts affected by disruptive factors. Specifically, we propose UGGC and UEMA modules for the stage-2 network. They can integrate uncertainty information and refine features during the encoding and decoding stages, respectively.

\begin{figure}[t]
    \centering 
    \includegraphics[width=0.7\textwidth]{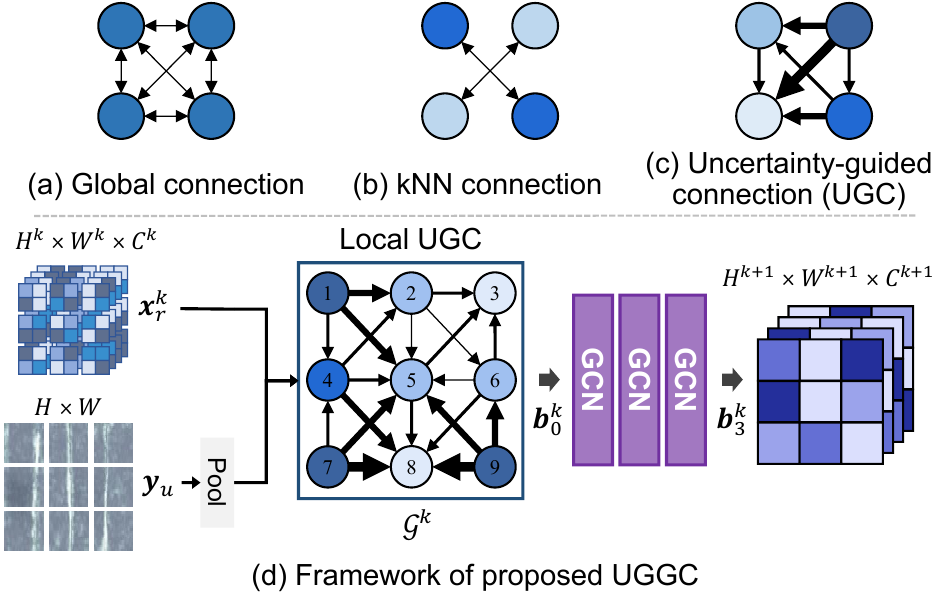}
    \caption{\textit{Top:} Illustration of three edge connection strategies, including (a) global connection, (b) k-nearest neighbor connection, and our (c) Uncertainty-Guided Connection (UGC). \textit{Bottom:} (d) Framework of the proposed UGGC based on the local UGC.} 
	\label{fig:uggc} 
\end{figure}

\label{sec:refine}
\paragraph{UGGC: Uncertainty-Guided Graph Convolution.} 
Self-attention techniques have been widely studied in media forensics \cite{span,mvss,trufor,emt,transforensics,add}, leveraging global information to distinguish manipulated from pristine regions \cite{pscc}. However, the prevalence of disruptive factors in scientific images complicates this task, potentially causing erroneous predictions due to the transmission of unreliable information. This issue remains unresolved even with recent k-nearest neighbor (kNN) connection strategies \cite{vig}.
To address this, we propose the Uncertainty-Guided Graph Convolution (UGGC) module (see \figref{fig:uggc}(d)), which forms a local directed weighted graph from feature maps, explicitly guiding interactions among regions. It can be seen in Figs.~\ref{fig:uggc}(a)-(c), differing from global and kNN connections, we employ an Uncertainty-Guided Connection (UGC) strategy to control information flow direction and intensity between regions. In UGGC, we add local connection constraints to UGC, denoted as local UGC, to help capture subtle differences such as texture and artifact inconsistencies. 

In the specific pipeline of UGGC, we first divide the input feature maps $\boldsymbol x_r^k$ and uncertainty map $\boldsymbol y_u$ into patches of size $n_p \times n_p$, where $k=1,2,3,4$. 
To explicitly control the information flow among regions, we constructed the local directed weighted graph $\mathcal{G}^k(\mathcal{N}^k, \mathbf{A}^k)$ using the proposed local UGC strategy. 
The node set $\mathcal{N}^k$ contains $H_n^k \times W_n^k = H^k / n_p \times W^k / n_p$ nodes, each node representing a corresponding patch of $\boldsymbol x_r^k$. The feature value of each node is the average pooled feature value of the corresponding patch. The number of channels for each node is consistent with $\boldsymbol x_r^k$. We use $\boldsymbol b_{0}^k\in \mathbb{R}^{(H_n^k \times W_n^k) \times C^k}$ to represent the initial node features of $\mathcal{N}^k$.

To construct the edge set $\mathbf{A}^k$, we first calculate an initial directed weighted adjacency matrix $\mathbf{E}^k$. The edge weight between nodes $\mathcal{N}^k_i$ and $\mathcal{N}^k_j$ is calculated as \eqnref{eq:edge1}.  
\begin{equation}
	\small
	\mathbf{E}^k_{i, j}=\left\{
	\begin{aligned}
		\boldsymbol y_u^k(j) -  \boldsymbol y_u^k(i) & , & \boldsymbol y_u^k(i) < \boldsymbol y_u^k(j), \\
		0 & , & \boldsymbol y_u^k(i) \geq \boldsymbol y_u^k(j),
	\end{aligned}
	\right.
	\label{eq:edge1}
\end{equation}
where $\boldsymbol y_u^k(\cdot)$ represents the average uncertainty of pixels within the corresponding patch in UGGC. As presented in \eqnref{eq:edge2}, $\mathbf A^k$ is derived by adding the local constraint to $\mathbf{E}^k$.
\begin{equation}
	\small
	\mathbf A^k_{i, j}=\left\{
	\begin{aligned}
		\mathbf{E}^k_{i, j} & , & \lvert i-j \rvert < 1 \textrm{ and } \lvert r(i)-r(j) \rvert < 1,\\
		0 & , & \textrm{otherwise}, 		
	\end{aligned}
	\right.
	\label{eq:edge2}
\end{equation}
where $r(\cdot)$ is the row index of the input node.
Following the construction of $\mathcal{G}_k$ are three serially connected Graph Convolution Networks (GCNs) \cite{gcn_iclr}. Under the local UGC constraints, the GCNs can regulate the information transmits from confident nodes to uncertain nodes. 
The specific calculation process is formulated as follows:
\begin{equation}
	\small
	\label{eq:gcn}
	\boldsymbol b_{\ell}^k = \tilde{\mathbf{D}^k}^{-\frac{1}{2}} \tilde{\mathbf{A}^k} \tilde{\mathbf{D}^k}^{-\frac{1}{2}} \boldsymbol b^k_{\ell-1} \mathbf{T}_{\ell},
\end{equation}
where $\boldsymbol b^k_{\ell} \in \mathbb{R}^{(H_n^k\times W_n^k) \times C_\ell^k}, \ell=1,2,3$ is the input feature map, $\tilde{\mathbf{A}^k}=\mathbf{A}^k+\mathbf{I}$ is the adjacency matrix added with an identity matrix $\mathbf{I}$, $\tilde{\mathbf{D}^k}$ denotes the diagonal node degree matrix of $\tilde{\mathbf{A}^k}$, and $\mathbf{T}_{\ell} \in \mathbb{R}^{C_{\ell-1}^k \times C_{\ell}^k}$ is a learnable weight matrix of the $\ell$-th GCN. To ensure the size and the number of channels of the output features from UGGC match those of the corresponding ResNet encoder, we let $H_n^k = H_{k+1}$, $W_n^k = W_{k+1}$, $C_1^k = C^k$, and $C_2^k=C_3^k=C^{k+1}$.

\paragraph{UEMA: Uncertainty-Enhanced Manipulation Attention.}
\begin{figure}[t]
	\centering 
	\includegraphics[width=0.7\textwidth]{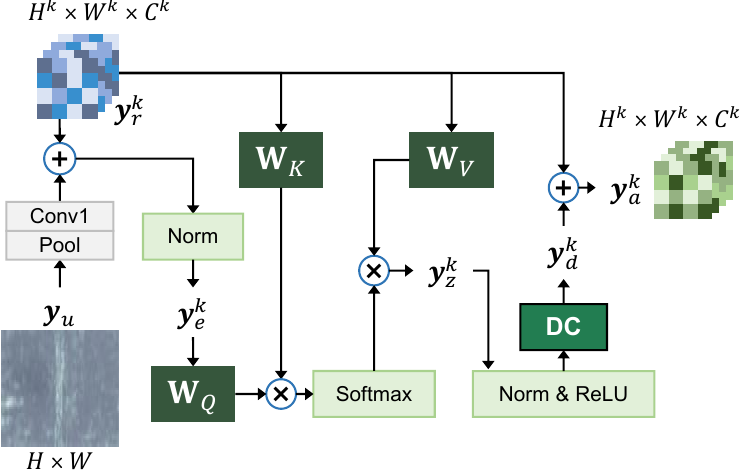}
	\caption{Framework of our Uncertainty-Enhanced Manipulation Attention (UEMA). ``DC'' is short for a depth-wise convolution layer.} 
	\label{fig:uema} 
\end{figure}
In the decoding phase, our goals are to direct more attention towards uncertain regions and align the features of spliced areas while ensuring their distinction from pristine areas.
To achieve these two goals simultaneously, we propose the UEMA module (see \figref{fig:uema}), which fully leverages the uncertainty map and robust features refined by UGGC to further enhance network performance. Due to the robust feature representation extracted by the UGGC, global correlation computations in UEMA during the decoding phase will not mislead the learning process.
We first use uncertainty maps to enhance feature maps and perform cross-attention on the feature maps before and after enhancement. Then, we adopt depth-wise convolution layers to adaptively adjust the weight of each channel to assist in refining. 

Taking $\boldsymbol y_r^{(k)}$ and $\boldsymbol y_u$ as the input of UEMA, we sequentially perform average pooling and a convolution layer with $1\!\times\!1$ kernels, making the size of $\boldsymbol y_u$ consistent with $\boldsymbol y_r^k$. Their output can be represented as \eqnref{eq:resize}.
\begin{equation}
	\small
	\boldsymbol y_e^k = \mathrm{BN}(\boldsymbol y_r^k + \mathrm{Conv1}(\mathrm{AveragePool}(\boldsymbol y_u))),
	\label{eq:resize}
\end{equation}
where $\mathrm{BN}(\cdot)$ is batch normalization, which is used to normalize the feature map enhanced by the uncertainty map. 

\begin{equation}
	\small
	\boldsymbol y_z^k = \mathrm{Softmax}(\frac{\mathbf W^k_Q(\boldsymbol y_e^k) \mathbf W^k_K(\boldsymbol y_r^k)^\intercal}{\sqrt{d^k}}) \mathbf W^k_V(\boldsymbol y_r^k),
\end{equation}
where $\mathbf W^k_Q(\cdot)$, $\mathbf W^k_K(\cdot)$, $\mathbf W^k_V(\cdot) \in \mathbb{R}^{d^k \times d^k}$ are the linear projections corresponding to the \textit{query}, \textit{key}, and \textit{value}, respectively. Here, $d^k = H^k \times W^k$ denotes the number of pixels in a single feature map. Using $\boldsymbol y_e^k$ as the \textit{query}, the UEMA can be essentially instructed to focus its attention on these areas of uncertain regions and regions suspected of splicing. In addition, $\boldsymbol y_r^k$ serves as the \textit{key} and \textit{value}, providing the original basis for global correlation computing.  
Lastly, acknowledging the different contributions from various channels towards the rectification of weakly predicted regions, we incorporate three cascaded depth-wise convolution layers $\mathrm{DC}(\cdot)$ (with kernel sizes of $3\!\times\!3$) \cite{mixformer} to further refine $\boldsymbol y_v^k$, allowing for an adaptive adjustment of weights among the channel dimension.
\begin{equation}
	\small
	\boldsymbol y_a^k=\boldsymbol y_z^k + \mathrm{DC}(\mathrm{BN}(\mathrm{ReLU}(\boldsymbol y_v^k))).
\end{equation}

\begin{figure}[t]
	\centering 
	\includegraphics[width=0.5\textwidth]{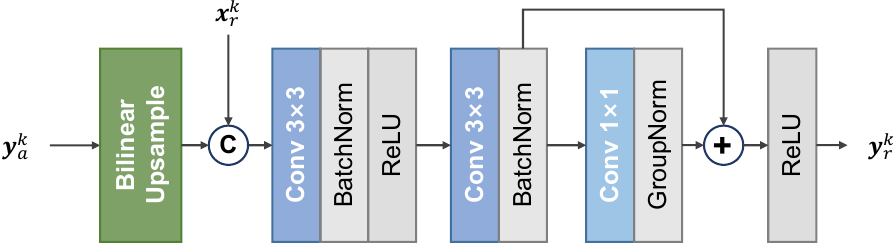}
	\caption{Structure of the decoder block in \model.}
	\label{fig:decoder}
\end{figure}

\subsection*{Implementation Details} 
We implement UGNet based on PyTorch 1.7.1. We train UGNet on a single RTX 3090 GPU for 200 epochs and optimize it using the Adam with an initial learning rate of $1 \times 10^{-4}$.
We set the batch size to 8 and resize all images and ground truth to $256 \times 256$. 
In Eqs. (7) and (8) of our manuscript, $\gamma_b$, $\gamma_d$, $\gamma_s$, and $\gamma_c$ are set to 0.7, 0.3, 1.0, and 0.35, respectively. 
In the stage-1 network, we sample $n_s\!=\!5$ times to estimate the uncertainty maps $\boldsymbol y_u$ and predict coarse masks $\boldsymbol y_m$. 
To improve the training efficiency of \model~and ensure the fairness of the ablation study, we first train the stage-1 network, then freeze its weights during the training phase of the stage-2 network. Besides, we use the weight of stage-1 to initialize stage-2 for better training efficiency.
The structure of our decoder is shown in \figref{fig:decoder}. The patch size $n_p$ is set to 2. In \model, the size of each intermediate feature map denoted as $(H^1, W^1, C^1)$, $(H^2, W^2, C^2)$, $(H^3, W^3, C^3)$, and $(H^4, W^4, C^4)$, is $(128, 128, 64)$, $(64, 64, 256)$, $(32, 32, 512)$, and $(16, 16, 1024)$, respectively. For the last UEMA module, we downsample the input feature maps and upsample the output ones (by 2$\times$) due to the limited GPU memory.

\subsection*{Details of Evaluation Metrics}
The F1 pays more attention to the number of correctly predicted spliced pixels, whereas MCC is frequently used for evaluating the performance of images with imbalanced pixels. MCC produces a high score only if the majority of the predicted negative and positive pixels are correct. These two pixel-level metrics are formulated as follows.
\begin{equation}
	\small
	F1= \frac{2\cdot TP}{2\cdot TP+FP+FN},
\end{equation}
\begin{equation}
	\small
	MCC\!=\!\frac{TP\!\cdot\!TN\!-\!FP\!\cdot\!FN}{\sqrt{(TP\!+\!FP)(TP\!+\!FN)(TN\!+\!FP)(TN\!+\!FN)}},
	%		MCC= \frac{TP\!\cdot\!TN\!-\!FP\!\cdot\!FN}{\sqrt{(TP\!+\!FP)\!\cdot\!(TP\!+\!FN)\!\cdot\!(TN\!+\!FP)\!\cdot\!(TN\!+\!FN)}}
\end{equation}
where $TP$, $TN$, $FP$, and $FN$ denote the number of true-positive, true-negative, false-positive, and false-negative pixel-level predictions, respectively. 

Since the ratio of spliced images to pristine images of our dataset protocols is set to 1$:$1, we adopt AUC and Acc which are suitable for evaluating performance when categories of datasets are balanced. AUC focuses on comprehensive performance under different decision thresholds, while accuracy excels at evaluating performance under a fixed threshold (default: 0.5). Acc is formulated as follows.
\begin{equation}
	\small
	Acc= \frac{TP_{cls} + TN_{cls}}{TP_{cls} + TN_{cls} + FP_{cls} + FN_{cls}},
\end{equation}
where 
$TP_{cls}$, $TN_{cls}$, $FP_{cls}$, and $FN_{cls}$ denote the number of images with true-positive, true-negative, false-positive, and false-negative image-level predictions, respectively.
We set the default decision thresholds of F1, MCC, and Acc to 0.5 for reasons that (1) in real-world applications, determining an optimal decision threshold for unseen testing data seems impractical, and (2) setting the same decision threshold for comparison is fairer.

\subsection*{Resource availability}
\subsubsection*{Lead contact}
% TODO: 记得改写
Further information and requests for resources should be directed to and will be fulfilled by the lead contact, Prof. Shuai Wang (wangshuai@buaa.edu.cn).

\subsubsection*{Materials availability}
This work did not provide new materials.
\subsubsection*{Data and code availability}
Existing datasets, \textit{i.e.,} Biofors and RSIIL, can be found at their official Github repositories (\url{https://github.com/vimal-isi-edu/BioFors} and \url{https://github.com/phillipecardenuto/rsiil}).

\noindent The source code of our URN is publicly available on Github (\url{https://github.com/lxbuaa/URN}). Our datasets SciSp can be downloaded from Zenodo (\url{https://zenodo.org/records/10989921}).
    
\section*{Acknowledgments}
This work was supported by the National Key Research and Development Program of China under Grant no.2022YFB3207700. The support funding was also from the National Natural Science Foundation of China under Grants 62272022 and U22A2009.

\section*{Author contributions}
Xun Lin was responsible for manuscript writing, code implementation, and main experiments.
Wenzhong Tang provided funding and performed theoretical analysis.
Yizhong Liu conducted ablation study and visualizations.
Haoran Wang performed robustness testing and contributed to manuscript revisions.
Yakun Ju conceptualized the details of UGGC and participated in manuscript revision.
Shuai Wang provided funding and conceptualized the overall structure.
Zitong Yu supervised the project and designed all experiments.

\section*{Declaration of interests}

The authors declare no competing interests.

\bibliography{references}

\end{document}